\documentclass{article} 
\usepackage{hyperref}
\usepackage{url}
\usepackage{amsfonts}
\usepackage{amsmath}
\usepackage[utf8]{inputenc}
\usepackage{microtype}
\usepackage{graphicx}
\usepackage{subfigure}
\usepackage{booktabs}
\usepackage{siunitx}

\usepackage{hyperref}

\usepackage[accepted]{icml2018}

\icmltitlerunning{Fast Parametric Learning with Activation Memorization}

\DeclareMathOperator*{\argmax}{arg\,max}

\newcommand{\model}{\hbox{Hebbian Softmax }}
\newcommand{\shortmodel}{\hbox{Hebbian }}

\mathchardef\mhyphen="2D

\def\argmax{\operatornamewithlimits{arg\,max}}

\def\c{\mathbf{c}}

\def\h{\mathbf{h}}

\def\p{\mathbf{p}}

\def\y{\mathbf{y}}

\def\0{\mathbf{0}}
\def\1{\mathbf{1}}
\def\2{\mathbf{2}}

\def\Ncal{\mathcal{N}}

\begin{document}
\twocolumn[
\icmltitle{Fast Parametric Learning with Activation Memorization}

\begin{icmlauthorlist}
\icmlauthor{Jack W Rae}{dm,ucl}
\icmlauthor{Chris Dyer}{dm}
\icmlauthor{Peter Dayan}{gatsby}
\icmlauthor{Timothy P Lillicrap}{dm,ucl}
\end{icmlauthorlist}

\icmlaffiliation{dm}{DeepMind, London, UK}
\icmlaffiliation{ucl}{CoMPLEX, Computer Science, University College London, London, UK}
\icmlaffiliation{gatsby}{Gatsby Computational Neuroscience Unit, University College London, UK}

\icmlcorrespondingauthor{Jack W Rae}{jwrae@google.com}
\icmlcorrespondingauthor{Timothy P Lillicrap}{countzero@google.com}

\icmlkeywords{Language Modeling, Softmax, Hebbian, Classification, Machine Learning, ICML}

\vskip 0.3in
]

\printAffiliationsAndNotice{}

\begin{abstract}
Neural networks trained with backpropagation often struggle to identify classes that have been observed a small number of times. In applications where most class labels are rare, such as language modelling, this can become a performance bottleneck. One potential remedy is to augment the network with a fast-learning non-parametric model which stores recent activations and class labels into an external memory. We explore a simplified architecture where we treat a subset of the model parameters as fast memory stores. This can help retain information over longer time intervals than a traditional memory, and does not require additional space or compute.
In the case of image classification, we display faster binding of novel classes on an Omniglot image curriculum task.
We also show improved performance for word-based language models on news reports (GigaWord), books (Project Gutenberg) and Wikipedia articles (WikiText-103) --- the latter achieving a state-of-the-art perplexity of $29.2$.

\end{abstract}

\section{Introduction}
Neural networks can be trained to classify discrete outputs by appending a softmax output layer. This is a linear map projecting the $d$-dimensional hidden output of the network to $m$ outputs, where $m$ is the number of distinct classes. A softmax operator \citep{bridle1990training} is then applied to produce a probability distribution over classes. The parameters in this softmax layer are typically optimized with the network's parameters by gradient descent. 

We can think of the weights in the softmax layer $\theta \in \mathbb{R}^{m \times d}$ as a set of $m$ vectors $\theta[i]; \; i = 1, \ldots, m$ that each correspond to a given class. When trained with a supervised loss, such as cross-entropy, each step of gradient descent pulls the parameter $\theta[y]$, corresponding to the class label $y$, towards having a greater inner product with the network output $h$, and pushes all other parameters $\theta[j]\; , \, j \neq y$ towards having a smaller inner product with $h$.

One shortcoming of neural network classifiers trained with backpropagation is that they require many input examples for a given class in order to predict it with reasonable accuracy. That is, many positive class examples and optimization steps are required to pull $\theta[i]$ towards a point in space where class $i$ can then be recognized. While the learner will have many opportunities to organize $\theta[i]$ parameters associated with frequent classes, infrequent class parameters will be poorly estimated. In domains where new classes are frequently introduced, or large-scale classification problems where some classes are very infrequently observed, this estimation problem is potentially quite serious.

One approach to speed up learning, which has received revived interest, is meta-learning. Here, meta-learning refers to algorithms which learn to produce  or manipulate learning algorithms \citep{thrun1998lifelong, hochreiter2001learning}, and it operates by learning over a distribution of tasks or datasets. A meta-learner applies knowledge from the global distribution of tasks to produce or optimize algorithms which specialize to a given task instance. Meta-learning of neural networks has seen promising results for applications such as parameter optimization \citep{andrychowicz2016learning, finn2017model} and classification \citep{santoro2016one, vinyals2016matching, zhou2018deep}. For classification, the networks are augmented with a differentiable external memory, and are trained with many rounds of data --- with class labels permuted between episodes.

Meta-learning can be very powerful for few-shot learning in cases where there is a set of similar prior data to meta-learn over, however it may not be practical for standalone datasets. For example, if one wants to model the grammar of computer code, it is unclear that a meta-learning system trained over natural language will be useful. Also memory-based meta-learning requires backpropagating from the read time to the original write time, which is not well suited to applications where writes and reads are separated by long time steps of conditional computation. In the case of modelling language, for example, infrequent words will not occur for large time intervals --- rendering memory-based meta-learning challenging.

The task of statistical language modelling itself is interesting to investigate issues of binding new or infrequent classes, because most classes (words) are infrequent \citep{zipf1935psychology} and new classes naturally emerge over time. Recent approaches to improve neural language models have involved augmenting the network with a non-parametric cache, which stores past hidden activations $h_{t-n}, \ldots , h_{t-1}$ and corresponding labels, $y_{t-n}, \ldots, y_{t-1}$ \citep{vinyals2015pointer, merity2016pointer, grave2016improving, kawakami2017learning, grave2017unbounded}. Attention over this cache provides better modelling of infrequent words that occur in a recent context, including previously unknown words \citep{gulcehre2016pointing}. However there is a diminishing return to increasing the cache size \citep{grave2016improving}, and once rare words fall outside the recent context the boost in predictive performance expires.

\begin{figure}
    \centering
    \includegraphics[width=0.4\textwidth]{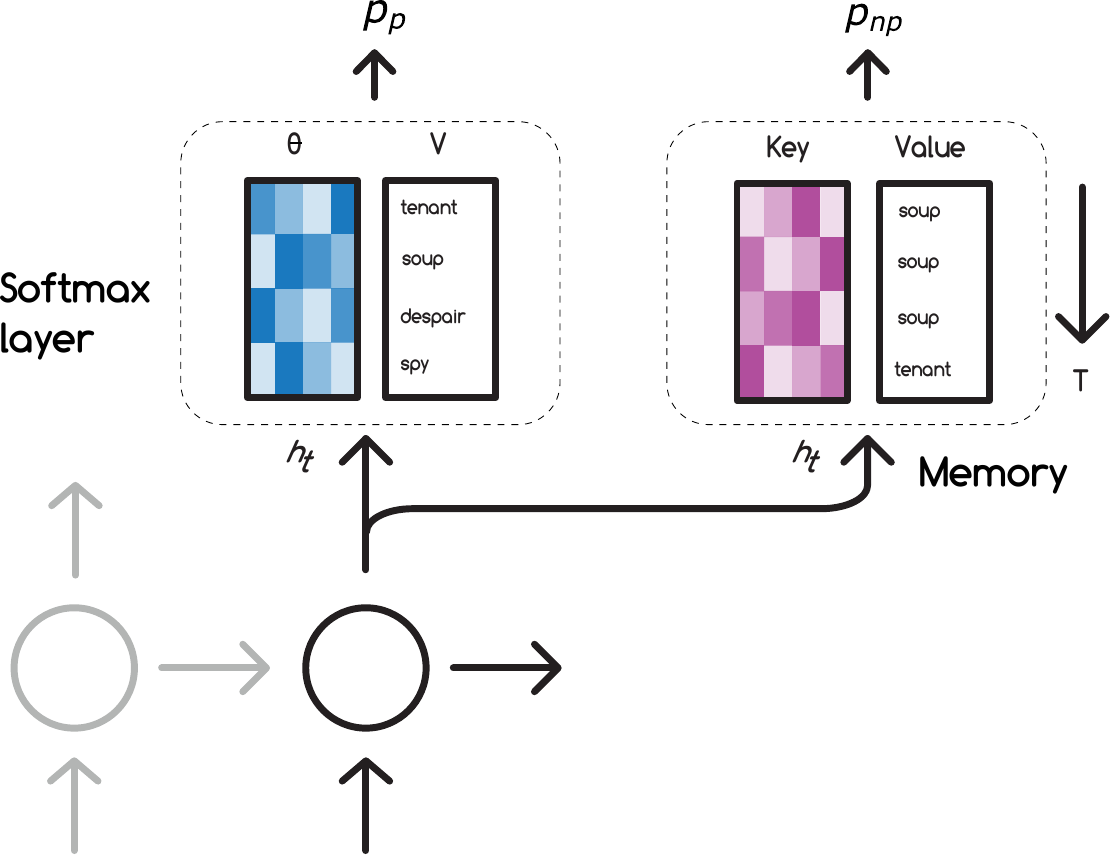}
    \caption{Mixture model of parametric and non-parametric classifiers connected to a recurrent language model. The non-parametric model (right hand side) stores a history of past activations and associated labels as key, value pairs. The parametric model (left hand side) contains learnable parameters $\theta$ for each class in the output vocabulary $V$. We can view both components as key, value memories --- one slow-moving, optimized with gradient descent, and one rapidly updating but ephemeral.}
    \label{fig:mixture}
\end{figure}

Motivated from these memory systems, we explore a very simple optimization procedure where the network accumulates activations $h_t$ directly into the softmax layer weights $\theta[y_t]$ when a class $y_t$ has been seen a small number of times, and uses gradient descent otherwise. Accumulating or smoothing network activations into the weights actually corresponds to the well-known Hebbian learning update rule $W[i, j] \leftarrow \frac{1}{n} \sum_{t = 1}^n x_t^i x_t^j$  \citep{hebb1949organization} in the special case of classification on the output layer, where $W, x_t^i, x_t^j$ correspond to $\theta, h_t, y_t$ respectively. We see that mixing the two rules provides better initial representations and can also preserve these representations for much longer time spans. This is because memorized activations for one class are not competing for space with activations from other (more frequent, say) classes --- unlike a conventional external memory. In this sense, the parameters become an instance of a quickly updated compressed memory, we explore this idea in Section \ref{sec:compressed_memory}

We demonstrate this model adapts quickly to novel classes in a simple image classification task using handwritten characters from Omniglot \cite{lake2015human}. We then show it improves overall test perplexity for two medium-scale language modelling corpora, WikiText103 (wikipedia articles) \citep{merity2016pointer} and Project Gutenberg  (books) \footnote{Project Gutenberg. (n.d.). Retrieved January 2, 2018, from www.gutenberg.org} alongside a large-scale corpus GigaWord v5 (news articles) \citep{parker2011english}. By splitting accuracy over word frequency buckets, we see improved perplexity for less frequent words.

\section{Background}


\subsection{Memory}

There has been recent interest in models which store past hidden activations through time $h_1, h_2, \ldots, h_{t-1}$ into a memory matrix and query the contents with a differentiable attention mechanism. This has been applied to  machine translation \citep{bahdanau2014neural}, program induction \citep{graves2014neural, graves2016hybrid}, and question answering \citep{sukhbaatar2015end}. Memory-augmented neural networks have also been successfully applied to language modelling \citep{vinyals2015pointer, kawakami2017learning, merity2016pointer, grave2016improving, grave2017unbounded} to facilitate the learning of unknown words, capture the tendency for globally rare words to be repeated in close proximity, and to quickly adapt the network to contextually relevant prior text \citep{sprechmann2018memorybased}.

There are many variants of how to read from memory and mix this information with the network computations. One approach is to retrieve hidden activations and mix these with network activations in latent space \citep{gulcehre2016pointing}. Another approach is a classic mixture model, as shown in Figure \ref{fig:mixture}; the output probability distribution can be obtained by interpolating the probabilities $p_p, p_{np}$ from the parametric model and memory respectively.

For intuition we briefly explain a particular architecture, the Neural Cache \citep{grave2016improving}, whose operation is related to our model. The cache is a store of the last $n$ hidden activations along with their corresponding target output (next word) from a trained parametric language model, such as the Long Short Term Memory (LSTM) \citep{hochreiter1997long}. The conditional probability of a word $w$ occurring is proportional to the sum over kernalized inner product similarities between the current hidden state $h_t$ and past hidden states when word $w$ occurred.
\begin{equation}
\label{eq:cache}
p_c(w \mid h_t) \propto \sum_{i = t - n}^{t - 1} e^{h_t^T h_i} \, \mathbb{I}\{y_i = w\}
\end{equation}
Where $\mathbb{I}\{p\} = 1$ if $p$ is true, $0$ otherwise. This is then interpolated with the parametric language model using a fixed hyper-parameter, swept over during validation. Although the cache is of fixed size $n$, it can be defined to be very large with sparse attention and efficient data-structures \cite{rae2016scaling, kaiser2017learning, grave2017unbounded}.

\subsection{Language modelling}
We can model a sequence of text as the product of conditional word probabilities,
\[ p(w_{1}, w_2, \ldots, w_t) = \prod_{i = 1}^t p(w_i \mid w_1, w_2, \ldots , w_{i-1}) \]
which are estimated separately. Traditional $n$-gram models take frequency-based estimates of these conditional probabilities with truncated contexts $p_n = p(w_i \mid w_{i-n}, \ldots, w_{i-1})$ and smooth between them to estimate the full conditional probability, $p(w_i \mid w_1, \ldots, w_{i-1}) = \sum_{j=1}^n \lambda_j p_j$. A popular approach is Kneser-Ney smoothing \citep{kneser1995improved}. More recently, neural language models such as LSTMs and convolutional neural networks directly model the conditional probabilities through sequence-to-sequence training and achieve state-of-the-art performance in many established benchmarks \citep{collobert2008unified, sundermeyer2012lstm, kalchbrenner2014convolutional, jozefowicz2016exploring, dauphin2016language, melis2017state}.

\section{Model}
\label{sec:model}

We propose \model \!, a modification of the traditional softmax layer with an updated learning rule. \model contains the same linear map from the hidden state to the output vocabulary, but learns by smoothing hidden activations into the weight parameters for novel classes whilst concurrently applying gradient descent. This is to facilitate faster binding of novel classes, and improve learning of infrequent classes. We note this corresponds to a learning rule that transitions from Hebbian learning to gradient descent, and we will show that the combination of the two learning rules works better than either one in isolation.

Many of the features of \model are motivated from memory systems, and the theory of complementary learning systems in the brain \cite{mcclelland1995there}. During training, the weights corresponding to a given class will initially correspond to a compressed\footnote{The memory is compressed because multiple activations corresponding to the same class are smoothed into one vector, instead of being stored separately.} episodic memory store --- with new activations memorized and older activations eventually forgotten. 

The parameters of the softmax layer are treated both as regular slow-adapting network parameters through which gradients flow to the rest of the network, and fast-adapting memory slots which are updated sparsely without altering the rest of the network. In comparison to an external memory, the advantage of \model is that it is simple to implement and requires almost no additional space or computation.

We will describe the learning rule in detail, and contrast the conditional probabilities from \model to those generated by a non-parametric cache. We also generalize the memorization procedure in Section \ref{sec:alternates} as an instance of a secondary fast-learning overfitting procedure with respect to a euclidean objective, and explore several promising variant objective functions. 
\subsection{Update Rule}
\begin{figure}
    \centering
    \includegraphics[width = 0.4 \textwidth]{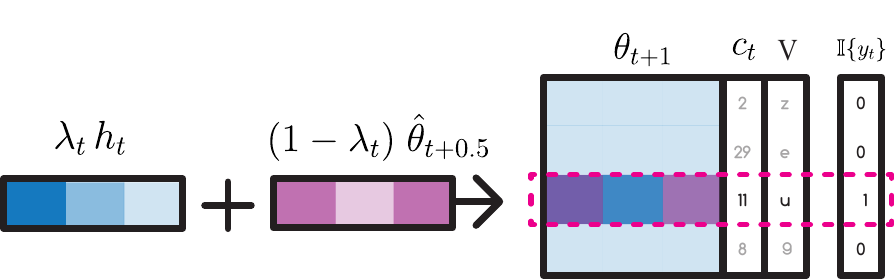}
    \caption{Update rule. Here the vector $\hat\theta_{t+0.5}$ denotes the parameters $\theta_t[y_t]$ of the final layer softmax corresponding to the active class $y_t$ after one step of gradient descent. This is interpolated with the hidden activation at the time of class occurrence, $h_t$. The remaining parameters are optimized with gradient descent. Here, $\mathbb{I}\{y_t\}$ is the one-hot target vector, $V$ denotes the vocabulary of classes, and $c_t$ is defined to be a counter of class occurrences during training --- which is used to anneal $\lambda_t$ as described in \eqref{eq:anneal}. }
    \label{fig:update}
\end{figure}
Given the weights of a linear projection $\theta \in \mathbb{R}^{d \times m}$ in the final softmax layer of a network, we calculate the gradient descent update with respect to a cross-entropy loss,
\begin{align}
    \label{eq:sgd_update}
    \hat\theta_{t + 0.5}[i] \leftarrow
    \begin{cases}
        \theta_t[i] - \alpha \, (p_i - 1) \, h_t & i = y_t \\
        \theta_t[i] - \alpha \, p_i \, h_t & i \neq y_t
    \end{cases}
\end{align}
where $p_i = e^{h_t^T \theta_i} / \sum_{j=1}^{n} e^{h_t^T \theta_j}$ is the probability output from the softmax, and $\alpha$ is the learning rate. In practice the gradient descent update $\hat\theta_{t + 0.5}$ can be calculated with adaptive optimizers, such as RMSProp \citep{tieleman2012lecture}. This is interpolated with the previous layer's hidden activation $h_t$ for the active class $y_t$,
\begin{align}
\label{eq:update_rule_small}
\theta_{t+1}[i] \leftarrow 
    \begin{cases}
        \lambda_t \, h_t + (1 - \lambda_t ) \, \hat\theta_{t+0.5}[i] & i = y_t \\
        \hat\theta_{t+0.5}[i] & i \neq y_t \; ,
    \end{cases}
\end{align}
as illustrated in Figure \ref{fig:update}. When $\lambda_t = 1$ this corresponds to the rule $\theta_{t+1} \leftarrow h_t \cdot \mathbb{I}\{y_t\}$ where $\mathbb{I}\{y_t\} \in [0, 1]^m$ is a one-hot target vector. In this case Hebbian update rule, $W_{ij} \leftarrow x_i x_j$ for $x_i = h_t$ the hidden output and $x_j = \mathbb{I}\{y_t\}$ the target. Naturally when $\lambda = 0$ this is gradient descent, and so we see \model is mixture of the two learning rules. All remaining parameters in the model are optimized with gradient descent as usual. 

When mixing the two learning rules, we would like to benefit from fast initial learning of classes that have not been seen many times, along with stable consolidation of frequently seen classes. As such we do not want $\lambda_t$ to be constant, but instead something that is eventually annealed to zero. We add an additional counter array $\c \in \mathbb{Z}^m$ which counts class occurrences, and propose an annealing function of
\begin{equation}
\label{eq:anneal}
\lambda_t = \max(1 \, /\, \c[y_t], \; \gamma) \, \cdot \, \mathbb{I}\{\c[y_t] < T\}
\end{equation}
where $\gamma,\, T$ are tuning parameters. $T$ is the number of class occurrences before switching completely to gradient descent and $\gamma$ is the minimum activation mixing parameter. Although heuristic, we found this worked well in practice vs. a constant $\lambda$ or pure annealing $\lambda_t = 1 / c[y_t]$. If training from scratch, we suggest setting $\gamma = 1 / N_{min}$ and $T = N_{min} \times (\hbox{\# epochs until convergence})$ where $N_{min}$ is the minimum number of occurrences of any class in a training epoch. This is to ensure we smooth over many class examples in a given epoch, and the memorization of activations continues until the representation of $h_t$ stabilizes. We describe the full algorithm in Algorithm \ref{alg:update}, including details for training with minibatches.
\begin{algorithm}[tb]
   \caption{\model batched update}
   \label{alg:update}
\begin{algorithmic}
   \STATE --- At iteration $0$
   \STATE $\gamma \leftarrow $ min. discount (hyper-parameter)
   \STATE $T \leftarrow $ smoothing limit (hyper-parameter)
   \STATE $M \leftarrow $ num. classes
   \STATE $B \leftarrow $ batch size
   \STATE $\c_{0}[i] \leftarrow 0; \quad i = 1, \ldots, M$
   \STATE --- At iteration $t$
    \STATE $\h_{t, 1:B} \leftarrow $ softmax inputs
    \STATE $\p_{t, 1:B} \leftarrow $ softmax outputs
    \STATE $\y_{t, 1:B} \leftarrow $ target labels
    \STATE $\hat\theta_{t + 0.5} \leftarrow $\texttt{SGD}$(\theta_t, \h_{t, 1:B}, \p_{t, 1:B}, \y_{1:B})$
    \FOR{$i = 1, \ldots, M$}
        \STATE $n_{t,i} \leftarrow \sum_{j=1}^B \, \mathbb{I}\{y_{t, j}=i\}$ 
        \IF{$n_{t, i} > 0$}
            \STATE $\lambda_{t, i} \leftarrow \max(1 / \c_t[i], \gamma) \, \mathbb{I}\{\c_t[i] < T \} $
            \STATE $\bar{h}_{t, i} \leftarrow \frac{1}{n_{t, i}} \sum_{j=1}^B h_{t, j} \mathbb{I}\{y_{t, j}=i\} $
            \STATE $\theta_{t+1} \leftarrow \lambda_{t, i} \bar{h}_{t, i} + (1 - \lambda_{t, i}) \hat\theta_{t+0.5}[i]$
        \ELSE
            \STATE $\theta_{t+1} \leftarrow \hat\theta_{t + 0.5}[i]$
        \ENDIF
        \STATE $\c_{t+1}[i] \leftarrow \c_t[i] + n_{t, i}$
    \ENDFOR

\end{algorithmic}
\end{algorithm}

The final layer trains with a two-speed dynamic. For some training steps the full network will be optimized slowly via gradient descent as usual (when frequently-encountered classes are observed), and for other time steps a sparse subset of parameters will rapidly change. The remaining network parameters are optimized with gradient descent.

It is worth noting that simply increasing the learning rate of the softmax layer, or running multiple steps of optimization on rare class inputs, would not achieve the same effect. The value $\theta[y_t]$ would indeed be pulled towards a large inner product with $h_t$, however neighbouring parameters $\theta[i] ; \; i \neq y_t$ would be pushed towards a large negative inner product with $h_t$ and this could lead to catastrophic forgetting of previously consolidated classes. Instead we allow gradient descent to slowly push neighbouring parameters away, and thus disambiguate similar classes in a gradual fashion.

\subsection{Relation to cache models}
\label{sec:compressed_memory}
We can consider the weights constructed from the above optimization procedure as a compressed memory storing historic activations. We contrast the output probabilities of \model with those produced from a non-parametric cache model.

Recall the conditional probability of a class, $w$, given a cache of previous activations \eqref{eq:cache}. If we set $I_w(j)$ to be the time step of j-th most recent occurrence of $w$, then we can re-write the cache probability,
\begin{align}
\label{eq:cache_rewrite}
    p_c(w \mid h_t) & \propto \displaystyle\sum_{i = t - n}^{t - 1} e^{h_t^T h_i} \mathbb{I}\{y_i = w\} \notag \\
    & = \displaystyle\sum_{j = 1}^{N_w} e^{ g(j) \, h_t^T h_{I_w(j)}}
\end{align}
where $g(j) = -\infty \hbox{ if } j < t - n \hbox { and } 1 $ otherwise, is a weighting function which places uniform weight to the attention over classes in the past $n$ time steps. However if we wish to characterize infrequent classes, we may want a weighting scheme with a larger time horizon that has a smooth decay.

If we modified the cache to have infinite memory capacity and used a geometric weighting scheme to decay the contribution of the $j$-th most recent activation corresponding to the given class, e.g. $g(j) = \lambda \, (1 - \lambda)^{j-1}$, then the resulting conditional probability is,
\begin{equation}
\label{eq:cache_infinite}
\tilde{p}_c (w \mid h_t) \propto \sum_{j = 1}^{N_w} e^{\lambda \, (1 - \lambda)^{j-1} \, h_t^T h_{I_w(j)}}
\end{equation}
where $N_w$ is the total number of occurrences of class $w$. Let us now consider the conditional probability from \model for class $w$, where $w$ has been observed less than $T$ times. If $\theta$ has not received large gradients from the occurrence of nearby neighboring classes, and we fix $\lambda_t = \lambda$ over time, then (\ref{eq:update_rule_small}) gives
\[\theta_i \approx \sum_{j = 1}^{N_w}  \lambda \, (1 - \lambda)^{j-1} h_{I_w(j)} \; ,\]
plugging this into our softmax conditional probability,
\begin{align*}
p_{\theta}(w \mid h_t) \propto e^{h_t^T \theta_w} & \approx e^{h_t^T \sum_{j=1}^{N_w} \lambda \, (1 - \lambda)^{j-1} h_{I_w(j)}} \\
& = \prod_{j=1}^{N_w} e^{\lambda \, (1 - \lambda)^{j-1} h_t^T h_{I_w(j)}} \, .
\end{align*}
we see the parametric \model actually becomes a proxy for the conditional probability output by the non-parametric infinite cache model $\tilde{p}_c$. Past activations now have a geometric contribution to the probability, versus the cache's arithmetic reduction \eqref{eq:cache_infinite}. This form is useful because we can compute $p_{sm}$ much more efficiently than $\tilde{p}_c$ and it does not require storing the entire history of past activations.

\subsection{Alternate Objective Functions}
\label{sec:alternates}
We briefly discuss a generalization of the \model update by casting it as an overfitting procedure to an inner objective function. Recall equation \eqref{eq:update_rule_small} for parameters corresponding to the active class, 
\begin{equation*}
\theta_{t+1}[i] \leftarrow \lambda_t \, h_t + (1 - \lambda_t) \, \hat\theta_{t+0.5}[i].
\end{equation*}
We can re-phrase this as smoothing $\hat\theta_{t+0.5}[i]$ with the trivial solution to a euclidean objective function, which we overfit to. 
\begin{align*}
& \theta_{t+1}[i] \leftarrow \lambda w^* + (1 - \lambda) \, \hat\theta_{t+0.5}[i]  \\
\vspace{0.5em} & w^* \leftarrow \argmax_w -||w - h_t||_2 
\end{align*}
From this perspective we are performing a two-level optimization procedure. The outer optimization loop is the mixture of gradient descent and exponential smoothing, and the inner optimization loop determines a good value for $w^*$ based on the activation $h_t$ and the current parameters.

We consider several other objective functions that are more expensive to compute, but may be preferable to a simple Euclidean distance. Notably, switching to inner product similarity (\texttt{IP}), and also incorporating a cost to parameter similarity (\texttt{SVM, Smax}) to push $w^*$ towards $h_t$ but away from neighbouring parameters --- to avoid confusion or interference with other classes. As we keep neighbouring parameters fixed, we hope to avoid the catastrophic forgetting typically associated with model overfitting. We list the set of objectives considered,
\begin{align}
\label{eq:overfitting_procedures}
& w^* \leftarrow \argmax_w \; g(w) \notag \\
& g_{\tiny \hbox{L2}}(w) = -||w - h_t||_2 \\
& g_{\tiny \hbox{IP}}(w) = w^T h_t \\
& g_{\tiny \hbox{SVM}}(w) = w^T h_t - \kern-1em \displaystyle\sum_{\theta_j \in \Ncal_k(h_t)} \kern-1em \xi \, w^T \theta_j \, \cdot \, \mathbb{I}(w^T \theta_j > \epsilon) \\
& g_{\tiny \hbox{Smax}}(w)  = e^{w^T h_t} / \displaystyle\sum_{\theta_j \in \Ncal_k(h_t)} e^{w^T \theta_j}
\end{align}
where $\Ncal_k(h_t)$ refers to the $k$ nearest parameters to the activation $h_t$ that do not correspond to $y_t$, the class label. Including all $M$ parameters in $\theta_t$ would make the inner optimization loop very slow, so we choose a sparse subset $k \ll M$. These are all optimized under the hard norm constraint $||w||_2 < 10$ with gradient descent for multiple steps, typically $20$, at a given point in training. 

\section{Results}
\subsection{Image Curriculum}

We apply \model to the problem of image classification. We create a simple curriculum task using Omniglot data \cite{lake2015human}, where a subset of classes ($30$) are initially provided, and $5$ new classes are added when test performance exceeds a threshold ($60\%$). Although this is a toy setup, it allows us to investigate the basic properties of fast class binding without other confounding factors, found in real-world problems.

Omniglot contains handwritten characters from $50$ alphabets, totalling $1,623$ unique character classes. There are $20$ examples per class. We partition the first $5$ examples per class to a test set, and assign the rest for training.

We use the same architectural setup as \textit{Matching Networks} \citep{vinyals2016matching} where the images are re-sized to $28 \times 28$ and a $4$ layer convolutional neural network is used. Each layer has $64$ filters, $3\times 3$ convolutions, batch normalization, ReLU activations, and $2 \times 2$ max pooling. Each channel maps the input to a scalar, so the resulting hidden size is $64$. All weight parameter in the softmax are initialized with Glorot initialization \citep{glorot2010understanding}. Models were trained with $20$\% dropout on the final layer and a small amount of data augmentation was applied to training examples (rotation $\in [-30, 30]$, translation) to avoid over-fitting. Otherwise the models quickly plateau on a low level. For the \model update, we store the pristine hidden activation pre-dropout. Unlike many one-shot Omniglot papers, we do not train in a meta-learning setup --- namely, labels are not shuffled between episodes.

We trained the convnet classifier with RMSProp and swept over learning rates $\alpha \in [1e-4, 4e-2]$ to find the fastest-learning baseline softmax model. A value of $\alpha = 8e-3$ was the largest learning rate to provide stable learning (see Figure \ref{fig:omniglot_lr} in Appendix \ref{appendix:omniglot}). We then compared the regular softmax layer with \model, both placed on the deep convnet.

\begin{figure}
    \centering
    \includegraphics[width=0.45 \textwidth]{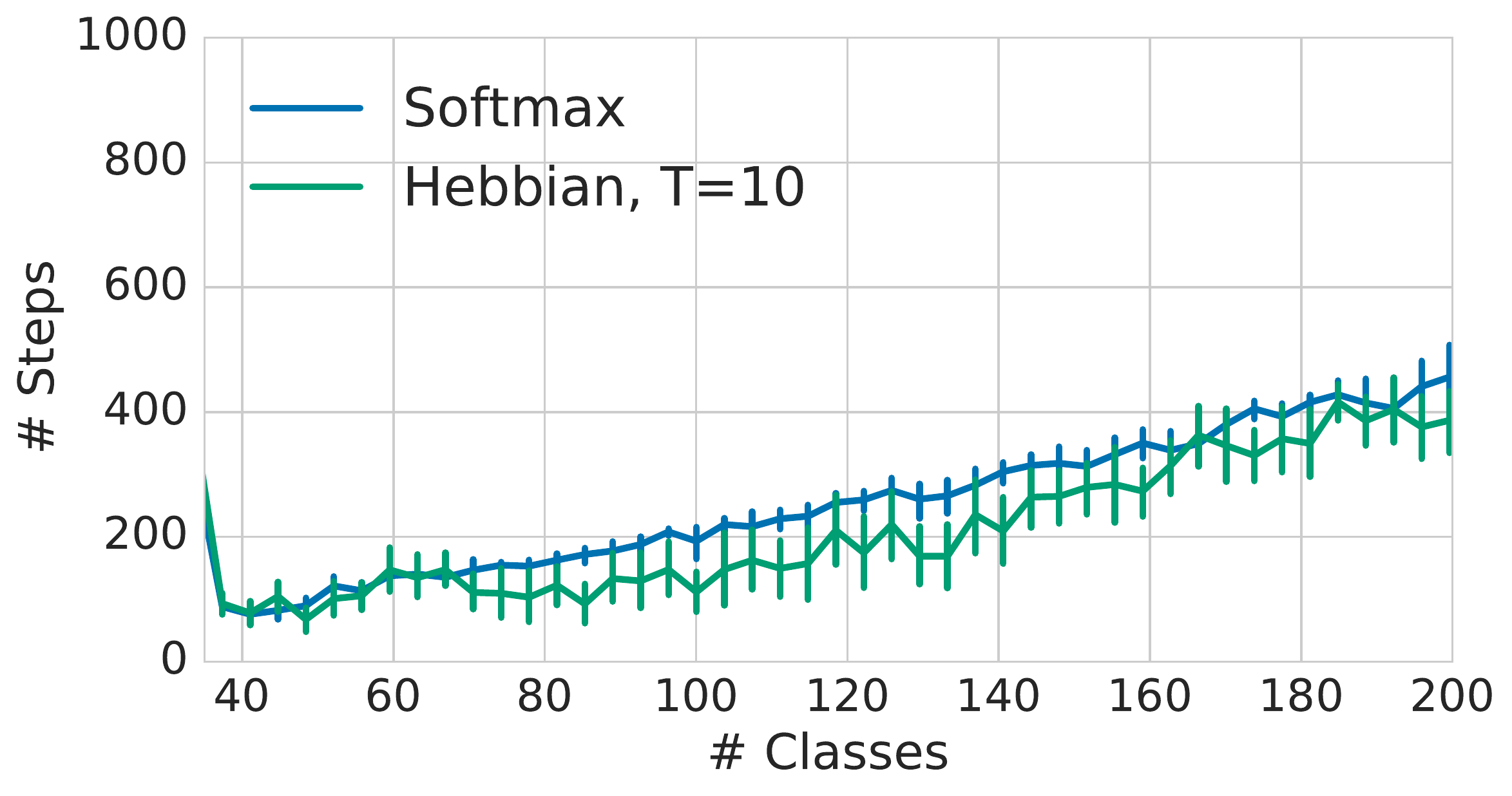}
    \caption{Number of training steps taken to complete each level on the Omniglot curriculum task. Comparisons between the \model and softmax baseline are averaged over $10$ independent seeds. As classes are sampled uniformly, we expect the number of steps taken to level completion to rise linearly with the number of classes.}
    \label{fig:omniglot_levels}
\end{figure}

If we inspect the number of steps spent on each level averaged over $10$ seeds, we see in Figure \ref{fig:omniglot_levels} that the model is noticeably more data efficient after $80$ total classes. Although it is still far from one-shot, there is a $1-2X$ data efficiency gain on average. In Appendix \ref{appendix:omniglot}, Figure \ref{fig:omniglot} we show the progression of the curriculum in terms of the number of classes shown versus training steps. \model progresses through the curriculum at identical speed to the softmax until around $1,000$ steps of training, from then on it begins to bind new classes and complete the each level faster.

\subsection{Language Modelling}
We would like to evaluate \model in the context of a large-scale classification task, where some classes are infrequently observed. Word-level language modelling is an ideal fit because it satisfies both criteria, and there are established performance benchmarks. Some large-scale language modelling corpora require the use of efficient softmax approximations, such as the adaptive softmax \citep{grave2016efficient} or hierarchical softmax \citep{goodman2001classes} due to the very large vocabulary size. To reduce confounding factors, we restrict ourselves to applications where the full softmax can be used. We investigate two medium-sized corpora, WikiText-103  which contains just over $100M$ tokens derived from Wikipedia articles \citep{merity2016pointer}, and Gutenberg which contains a subset of open-access texts from Project Gutenberg listed in Appendix \ref{appendix:gutenberg}. The idea is that Wikipedia articles should cover factual information, where the style of writing is somewhat consistent and named entities may appear across many articles; whereas books should be more self-contained (unique named entities) and stylistically different. We also consider a very large corpus, GigaWord v5, which is a collection of articles from eight press associations exceeding a decade's worth of global news.

\subsubsection{Model details}
For WikiText-103 we swept over LSTM hidden sizes $\{1024, 2048, 4096\}$, no. LSTM layers $\{1, 2\}$, embedding dropout $\{0, 0.1, 0.2, 0.3\}$, use of layer norm \citep{ba2016layer} $\{\textit{True}, \textit{False}\}$, and whether to share the input/output embedding parameters $\{\textit{True}, \textit{False}\}$ totalling $96$ parameters. A single-layer LSTM with $2048$ hidden units with tied embedding parameters and an input dropout rate of $0.3$ was selected, and we used this same model configuration for the other language corpora. We trained the models on $8$ P100 Nvidia GPUs by splitting the batch size into $8$ sub-batches, sending them to each GPU and summing the resulting gradients. The total batch size used was $512$ and a sequence length of $100$ was chosen. We did not pass the state of the LSTM between sequences during training, however the state is passed during evaluation. 

\begin{figure}[]
    \centering
    \includegraphics[width=0.45 \textwidth]{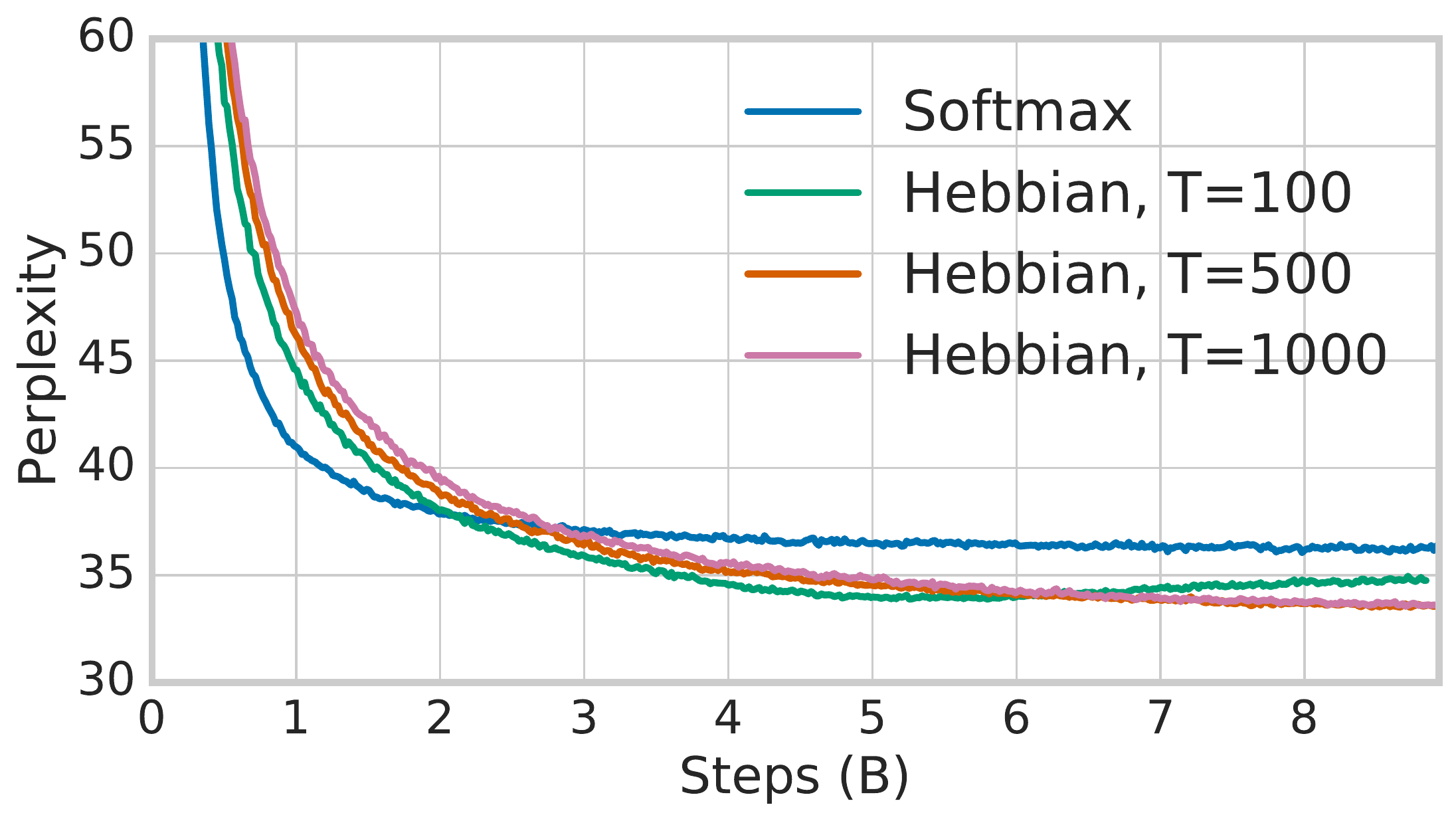}
    \caption{Validation perplexity for WikiText-103 over $9$ billion words of training ($\approx 90$ epochs).
    The LSTM drops to a perplexity of $36.4$ with a regular softmax layer, and $34.3$ with the \model, $T=500$, when representations from the LSTM begin to settle. For tuning parameter $T$; $T=100$ converges quicker, but begins to overfit after $5.5$B training words (coinciding when all classes have been observed at least $100$ times).}
    \label{fig:wiki_valid}
\end{figure}

\subsubsection{WikiText-103}
The WikiText-103 corpus contains $267,735$ unique words and each word occurs at least three times in the training set. We take the best LSTM parameter configuration (described above) as a baseline, and compare it to an identical model where the final layer is replaced with \model. We swept over the insertion limit parameter $T \in \{100, 500, 1000\}$ and discount factor $\gamma \in \{0.05, 0.1, 0.25\}$ using the validation set. We found $T = 500, \, \gamma = 0.25$ worked best, achieving a test perplexity of $34.3$ on this dataset (Table \ref{tab:wiki}). Inspecting the validation curves in Figure \ref{fig:wiki_valid} we see the \model initially hampers validation performance, until around $2$--$3$B training tokens have been consumed. This makes sense, as storing activations from prior layers of the network is only an effective strategy once the network has rich intermediate representations of its inputs. Inspecting Table \ref{tab:lm_breakdown} we see the test perplexity broken down by word frequency, the gain in overall performance is obtained from less frequent vocabulary.

We also investigate the model evaluated dynamically on the test using (a) a Neural Cache \citep{grave2016improving} and (b) Memory-based Parameter Adaptation (MbPA) \citep{sprechmann2018memorybased}. Hyper-parameter details for these models are detailed in Appendix \ref{appendix:wikitext}. The cache reduces the test perplexity by $1.6$ for the LSTM and $4.4$ for LSTM + \model. The addition of MbPA reaches a test perplexity of $29.2$ which is, to the authors' knowledge, state-of-the-art at time of writing.
\begin{table}
    \centering
    \caption{Validation and test perplexities on WikiText-103.}
    \begin{tabular}{lc c}
    \toprule
    & Valid. & Test \\
    \midrule
    LSTM \cite{graves2014neural} & - & 48.7 \\
    Temporal CNN \cite{bai2018convolutional} & - & 45.2 \\
    Gated CNN \cite{dauphin2016language} & - & 37.2 \\ 
    LSTM (ours) & 36.0 & 36.4 \\
    LSTM + Cache & 34.5 & 34.8 \\
    LSTM  + \shortmodel & 34.1 & 34.3 \\
    LSTM + \shortmodel + Cache & 29.7 & 29.9 \\ 
    LSTM + \shortmodel + Cache + MbPA & \textbf{29.0} & \textbf{29.2} \\
    \bottomrule
    \end{tabular}
    \label{tab:wiki}
\end{table}
\begin{table}[]
    \small
    \centering
    \begin{small}
    \caption{Test perplexity versus training word frequency. \model models less frequent words with better accuracy. Note the training set size of WikiText is smaller than Gutenberg, which is itself much smaller than GigaWord; so the $>10$K bucket includes an increasing number of unique words. This explains GigaWord's larger perplexity in this bucket. Furthermore there were no words observed $<100$ times within the GigaWord $250$K vocabulary. A random model would have a perplexity of $|V| \approx 2.5e5$ for all frequency buckets.}
    \end{small}
    \begin{small}
    \begin{sc}
    \setlength{\tabcolsep}{4pt}
    \begin{tabular}{l c c c c c}
    \toprule
     & $>10$K & $1$K$ \mhyphen 10$K & $100 \mhyphen 1$K & $<100$ & All \\ 
     \midrule
    \textbf{WikiText-103} & & & \\
    \scriptsize{softmax} & 12.1 & 2.2e2 & 1.2e3 & 9.7e3 & 36.4 \\ 
    \scriptsize{\model} & 12.1 & 1.8e2 & 7.6e2 & 5.2e3 & 34.3 \\ 
    \midrule
    \textbf{Gutenberg} & & & & \\
    \scriptsize{softmax} & 19.0 & 9.8e2 & 6.9e3 & 8.6e4 & 47.9 \\ 
    \scriptsize{\model} & 18.1 & 9.4e2 & 6.6e3 & 5.9e4 & 45.5 \\ 
    \midrule
    \textbf{GigaWord} & & & & \\
    \scriptsize{softmax} & 39.4 & 6.5e3 & 3.7e4 & - & 53.5  \\ 
    \scriptsize{\model} & 33.2 & 3.2e3 & 1.6e4 & - & 43.7 \\ 
    \bottomrule
    \end{tabular}
    \end{sc}
    \end{small}
    \label{tab:lm_breakdown}
    \vskip -0.1in
\end{table}
\subsubsection{Gutenberg}
Books provide several different linguistic challenges to articles. The style of writing is intentionally varied between authors, and named entities can be wholly fictional --- confined to a single text. We extract a subset of English-language books from the corpus, strip the Gutenberg headers and tokenize the text (Appendix \ref{appendix:preprocess}). We select a dataset of comparable size to WikiText-103; $2042$ books in total with $2017$ training books ($175,181,505$ tokens), $12$ validation books ($609,545$ tokens), and $13$ test books ($526,646$ tokens) --- see Appendix \ref{appendix:gutenberg} for full details. We select all words that occur at least five times in the training set, a total vocabulary of $242,621$ and map the remainder to an unk token.

We use the same LSTM hyper-parameters as those chosen from the wikipedia sweep, and compare against \model with $T=100, T=500$ and $\gamma = 0.1$. Figure \ref{fig:gutenberg_valid} in Appendix \ref{appendix:gutenberg} shows the validation performance after $15$B steps of training, equating to roughly $80$ epochs and $6$ days of training with $8$ P$100$s training synchronously. After approximately $4$B steps of training the softmax performance is surpassed, and this gap widens even up to $15$B steps to a gap of $2-3$ points in perplexity. Similar to WikiText-103, we see in Table \ref{tab:lm_breakdown} the gain in perplexity is more pronounced over less frequent words.

\subsection{GigaWord v5}
We evaluate \model on a large-scale language modelling corpus. GigaWord is interesting because it is a vast collection of news articles, and there is a natural temporal order. We pre-process the dataset (Appendix \ref{appendix:preprocess}), select all articles from $2000$--$2009$ for the training set, and test on all articles from $2010$. The total number of training tokens is $4.0$B and the total number of test tokens is $260$M. The total unique tokens (after pre-processing) for the training set reaches $6$M, however for parity with the other experiments we choose a vocabulary size of $250$K. We use the same LSTM hyper-parameters and \model hyper-parameters, and train the model for $6$B steps, after which the models plateau in evaluation performance. We observe a $9.8$-point drop in perplexity, from $53.5$ to $43.7$, illustrated in Table~\ref{tab:lm_breakdown}. 

\subsection{Alternate Objective Functions}
We test out some of the alternate inner objective functions described in Section \ref{sec:alternates}. These are described in \eqref{eq:overfitting_procedures}, and  the inner objective functions include \textit{Euclidean, Inner Product, SVM, (sparse) Softmax}. These could be applied to any of the described experiments, we chose the WikiText-103 language modelling task because it is more comparable to prior work.

Although more expressive objective functions appear promising, in practice we see (Figure \ref{fig:alternates} in Appendix \ref{appendix:alternates}) that validation performance is roughly equivalent between all inner objective functions. This suggests the network activation $h_t$ naturally do not land too close to other class parameters, and the norm of activations is not too large or small, in comparison to the model parameters $\theta$. The latter may be due to the use of layer normalization from the LSTM.

\section{Related Work}
Few-shot classification has been investigated in a meta-learning setup with a mixture model of a parametric neural network and a non-parametric memory \citep{santoro2016one, vinyals2016matching}. Here, a subset of classes are used with permuted labels per episode, activations are stored to memory, and gradients are passed through the memory. This allows the network to shape its activations to be conducive to accurate retrieval and classification. Here, we do not meta-learn the activations stored into network parameters and instead rely on their representation being rich enough from regular parametric training. We do this to avoid backpropagating through time to the point of writing to memory, as the parameters may contain memories stored millions of time steps ago in the case of rare words.

In natural language processing memory-augmented models have been shown to improve the modelling of unknown words and adaptation to new domains \citep{grave2016improving,merity2016pointer,kawakami2017learning}. However in these works the memory is typically small and models the recent past. During evaluation the test activations and corresponding labels are stored in memory, and the model is evaluated dynamically --- adapting to the test data on the fly. Whilst dynamic evaluation provides insights into domain transfer, it is limited in applicability as the model may not receive ground-truth labels when launched into production.

More recent work has investigated methods of memorizing and searching over the training set to enhance performance \citep{kaiser2017learning, grave2017unbounded, gu2017search}. These approaches typically require complex engineering to efficiently index this memory store. Part of the benefit of \model is implementation simplicity.

Prior literature on the softmax operator for language modelling computational efficiency \cite{chen2015strategies,grave2016efficient} or tricks such as smoothing across many softmax layers \cite{yang2017breaking}. However these do not focus on increasing the data-efficiency or faster learning of infrequent classes.

Other architectures have been considered for fast learning, such as the `fast weights' auto-associative memory \citep{ba2016using}. This focuses on fast adaptation to recent information that persists over a short window of time. The LEABRA architecture \citep{o1996leabra} contains a mixture of contrastive Hebbian learning (GENEREC) \citep{o1996biologically} and gradient descent for fast and slow learning, however this cognitively-inspired model has not been shown to scale to large-scale classification problems. 

\section{Discussion}
This paper explores one way in which we can achieve fast parametric learning in neural networks, and preserve this knowledge over time. We show that activation memorization is useful for vision in the binding of newly introduced classes, beating a well tuned adaptive learning rate optimizer, RMSProp. 

For language we show improvement in the modelling of text with an extensive vocabulary. In the latter we show the model beats a very strong LSTM benchmark on three stylistically different corpora, and achieves state of the art on WikiText-103. This is achieved with effectively no additional compute or memory resources. Breaking down perplexity over word frequency bucket, we see that less frequent words are better modelled, as hypothesized. We suggest that \model could be applied to any classification domain with infrequent classes, or non-stationary data. It may also be useful in quickly adapting a pre-trained classifier to a new task / set of classes --- however this is beyond the scope of our initial investigation.

It would also be interesting to explore activation memorization deeper within the network, and thus in more general scenarios to classification. In this case, there is no direct feedback from a ground-truth class label and the update rule would not necessarily be an instance of Hebbian learning. A natural first step would be to generalize the ideas to large-scale softmax operators that are internal to the network --- such as attention over a large memory.

\section*{Acknowledgements}

The authors would like to thank Gabor Melis, Greg Wayne, Oriol Vinyals, Pablo Sprechmann, Siddhant Jayakumar, Charles Blundell, Koray Kavukcuoglu, Shakir Mohamed, Adam Santoro, Phil Blunsom, Felix Hill, Angeliki Lazaridou, James Martens, Ozlem Aslan, Guillaume Desjardins, and Chloe Hillier for their comments and assistance during the course of this project. Peter Dayan is currently on a leave of absence at Uber Technologies; Uber was not involved in this study.

\newpage 

\bibliography{refs}
\bibliographystyle{icml2018}

\newpage
\newpage

\appendix

\onecolumn

\section*{Appendix}

\section{Language Modelling}

\subsection{WikiText-103}
\label{appendix:wikitext}

\subsubsection{Dynamic Evaluation Parameters}
For the Neural Cache, we swept over the hyper-parameters:
\begin{itemize}
\item Softmax inverse temperature: $\theta_{cache} \in \{$0.1, 0.2, 0.3$\}$
\item Cache output interpolation: $\lambda_{cache} \in \{$0.05, 0.1, 0.15, 0.2, 0.25, 0.3, 0.35$\}$
\item Cache size $n_{cache} \in \{$1000, 5000, 8000, 9000, 10000$\}$
\end{itemize}
and chose $\theta_{cache} = 0.3, \, \lambda_{cache} = 0.1, \, n_{cache} = 10000$ by sweeping over the validation set. 

For the mixture of Neural Cache and MbPA we swept over the same cache parameters, alongside:

\begin{itemize}
\item MbPA output interpolation: $\lambda_{mbpa} \in  \{$0.02, 0.04, 0.06, 0.08, 0.10$ \}$,
\item Number of neighbours retrieved from memory: $K \in \{512, 1024\}$,
\item Number of MbPA steps: $n_{mbpa} \in \{1, 2 \}$
\end{itemize}

and selected $\lambda_{mbpa} = 0.04, \lambda_{cache} = 0.1, \theta_{cache} = 0.3, K = 1024, n_{mbpa} = 1, n_{cache} = 10000$. We also selected the MbPA learning rate $\alpha_{lr} = 0.3$, and the L2-regularization $\beta_{mbpa} = 0.5$ on the MbPA-modified parameters. The memory size for MbPA was chosen to be equal to the cache size.

\subsection{Gutenberg}

\label{appendix:gutenberg}

\subsubsection{Splits}

We downloaded a subset of books (listed below) from Project Gutenberg on January 2, 2018 from a mirror site (\url{https://www.gutenberg.org/MIRRORS.ALL}). We selected $2042$ English-language books under the $/1$ subdirectory. Each book has a unique id, we shuffled the books and split out a reasonably sized train, validation and test set. The book ids are listed below for these splits.

\textbf{Test} ($13$ books, $526,646$ tokens):

\texttt{\tiny 11959, 12211, 10912, 11015, 12585, 10827, 10268, 11670, 126, 1064, 11774, 12505, 11931}

\textbf{Validation} ($12$ books, $609,545$ tokens):

\texttt{\tiny 11399,  10003,  1202,  12213,  11177,  12856,  10516,  11635,  12315,  11804,  11249,  11163}

\textbf{Train} ($2017$ books, $175,181,505$ tokens)

\texttt{\tiny 10000, 10064, 1019, 10358, 10482, 10598, 10675, 10787, 10864, 10929, 10, 11112, 11190, 11257, 11363, 11449, 11526, 11612, 11715, 11825, 11920, 11987, 1204, 12118, 12184, 12249, 12326, 12405, 12478, 12582, 12691, 12806, 12895, 10001, 10065, 101, 1035, 10483, 10599, 10676, 10788, 10865, 10930, 11007, 11113, 11191, 11258, 11364, 11451, 11527, 11613, 11716, 11826, 11921, 11988, 12050, 1211, 12185, 12252, 12327, 12406, 1247, 12583, 12692, 12807, 12896, 10002, 10066, 10201, 10363, 10489, 1059, 1067, 10789, 10867, 10931, 11008, 11114, 11192, 11259, 11365, 11452, 11528, 11614, 1171, 11827, 11922, 11989, 12051, 12121, 12186, 12253, 12328, 12409, 12486, 12584, 12696, 12808, 12897, 10067, 10202, 10365, 1048, 105, 10684, 1078, 10868, 10932, 11009, 11115, 11193, 11260, 11366, 11454, 1152, 11615, 1172, 11828, 11923, 1198, 12052, 12122, 12187, 12254, 12329, 1240, 1248, 12697, 12809, 12898, 10004, 10068, 1020, 10366, 10490, 10600, 10687, 10790, 10869, 10933, 11010, 11119, 11194, 11263, 11367, 11455, 11530, 11623, 11734, 11829, 11924, 11990, 12054, 12123, 12188, 12256, 1232, 12412, 12490, 12589, 12699, 1280, 12899, 10005, 10069, 10210, 10367, 10491, 10601, 1068, 10791, 10870, 10934, 11012, 11120, 11195, 11264, 11368, 11456, 11531, 11624, 11735, 1182, 11926, 11991, 12055, 12124, 12189, 12257, 12330, 12413, 12491, 1258, 1269, 12810, 128, 10006, 1006, 10211, 10368, 10493, 10602, 10690, 10792, 10871, 10935, 11013, 11121, 11196, 11265, 11369, 11459, 11533, 11625, 11736, 11830, 11929, 11992, 12056, 12125, 1218, 12259, 12333, 12414, 12498, 12590, 12811, 12900, 10007, 10070, 10212, 10369, 1049, 10603, 10691, 10793, 10872, 10936, 11014, 11122, 11197, 11266, 11370, 1145, 11534, 11626, 11737, 11831, 11930, 11993, 12057, 12126, 12190, 1225, 12336, 12415, 124, 12591, 12700, 12813, 12901, 10008, 10071, 10213, 1036, 104, 10605, 10692, 10794, 10873, 10937, 11123, 11198, 11267, 11371, 11460, 11537, 11632, 11738, 11832, 11994, 12058, 12127, 12191, 12261, 12337, 12416, 12504, 12592, 1270, 12814, 12902, 10009, 10072, 10214, 10370, 1050, 10606, 10693, 10795, 10874, 10938, 11016, 11124, 111, 11268, 11372, 11461, 11538, 11633, 1173, 11833, 11932, 11995, 12059, 12128, 12192, 12262, 12340, 12417, 12593, 1271, 12815, 12904, 10010, 10073, 10216, 10371, 10510, 10607, 10694, 10796, 10875, 10939, 11017, 11125, 11200, 11269, 11373, 11462, 11539, 11634, 11740, 11834, 11933, 11996, 12060, 12129, 12193, 12263, 12341, 12418, 12506, 12594, 12732, 12816, 12905, 10011, 1007, 10217, 10372, 10609, 10698, 10797, 10876, 10940, 11018, 11127, 11201, 11270, 11376, 11464, 1153, 11741, 11835, 11934, 11997, 12061, 1212, 12194, 12264, 12342, 12419, 12507, 12595, 12736, 12817, 1290, 10012, 10084, 10219, 10373, 10517, 10610, 10699, 10798, 10877, 10942, 11019, 11128, 11202, 11271, 11377, 11468, 11540, 11636, 11742, 11836, 11935, 11998, 12062, 12130, 12195, 12265, 12343, 1241, 1250, 12596, 12737, 12819, 12915, 10013, 10085, 1021, 10374, 10518, 10611, 1069, 10799, 10878, 10943, 11020, 1112, 11203, 11272, 11378, 11469, 11541, 11637, 11743, 11837, 11936, 11999, 12063, 12131, 12196, 12269, 12344, 12420, 12511, 1259, 12738, 1281, 12916, 10014, 10090, 10222, 10375, 10519, 10612, 106, 1079, 10879, 10944, 11021, 11130, 11204, 11273, 11379, 1146, 11542, 11638, 11745, 11838, 11937, 119, 12064, 12132, 12197, 12270, 12345, 12421, 12512, 125, 12739, 12821, 12917, 10015, 10091, 10224, 10376, 1051, 10613, 10700, 10800, 1087, 10945, 11028, 11136, 11210, 11274, 11382, 11470, 11543, 11647, 11746, 11839, 11938, 11, 12066, 12133, 12198, 12272, 12346, 12422, 12513, 12600, 12740, 12823, 1291, 10016, 10092, 10225, 10377, 10520, 10615, 10707, 10801, 10880, 10946, 11029, 11137, 11211, 11275, 11383, 11471, 11544, 11648, 11749, 1183, 11939, 12001, 12067, 12134, 12199, 12277, 12349, 12423, 12514, 12601, 12741, 12825, 12922, 10017, 10095, 10226, 10378, 10523, 10616, 10708, 10803, 10881, 10947, 11030, 11138, 11212, 11276, 11385, 11472, 11545, 11649, 1174, 11840, 11941, 12002, 12068, 12135, 121, 12278, 12350, 12424, 12515, 12611, 12742, 12826, 12923, 10018, 10096, 1022, 10379, 1052, 10617, 10709, 10804, 10882, 10948, 11031, 11140, 11213, 11277, 11386, 11473, 11546, 1164, 11753, 11841, 11942, 12004, 12069, 12136, 12200, 12279, 12351, 12425, 12516, 12614, 12743, 12827, 12924, 10019, 10097, 10234, 10380, 10538, 10618, 10712, 10805, 10883, 10949, 11032, 11141, 11214, 11278, 11387, 11474, 11548, 11651, 11754, 11842, 11943, 12006, 1206, 12137, 12201, 1227, 12352, 12426, 12517, 12617, 12744, 12828, 12925, 1001, 10098, 1024, 10381, 10543, 10619, 10714, 10806, 10884, 1094, 11033, 11142, 11215, 11279, 11388, 11475, 11549, 11652, 1175, 11843, 11944, 12007, 12071, 12138, 12202, 12280, 12353, 12427, 1251, 12618, 12745, 1282, 12926, 10020, 10099, 10266, 10382, 10544, 1061, 10715, 10807, 10885, 10950, 11034, 11143, 11216, 11280, 11389, 11476, 1154, 11653, 11761, 11844, 11945, 1200, 12073, 12139, 12203, 12281, 12354, 12428, 12521, 12619, 12746, 12830, 12928, 10022, 100, 10267, 10383, 10545, 10620, 10716, 10808, 10886, 10954, 1103, 11144, 11217, 11281, 1138, 11477, 11550, 11654, 11762, 11845, 11946, 12010, 12074, 12140, 12204, 12282, 12357, 12429, 12522, 1261, 12747, 12832, 12929, 10023, 10100, 10386, 10546, 10621, 10717, 1080, 10887, 10955, 11045, 11145, 11218, 11282, 11390, 11478, 11551, 11655, 11763, 11846, 11947, 12013, 12077, 12141, 12205, 12283, 12358, 1242, 12523, 12622, 1274, 12833, 12933, 10024, 10101, 1027, 10388, 10550, 10622, 10720, 10811, 10888, 10956, 11047, 11146, 11219, 11283, 11391, 11479, 11552, 11656, 11764, 11847, 11948, 12014, 12078, 12142, 12206, 12285, 12359, 12430, 12524, 12628, 12750, 12834, 12934, 10025, 10102, 1028, 10389, 10551, 10623, 10721, 10812, 10889, 10957, 11050, 11147, 1121, 11284, 11392, 1147, 11553, 11658, 11765, 11848, 11949, 12015, 12079, 12143, 12207, 12286, 1235, 12431, 12525, 12629, 12753, 12835, 12935, 10029, 10103, 10291, 10392, 10554, 10624, 10722, 10813, 1088, 10958, 11051, 11148, 11221, 11289, 11395, 11480, 11554, 11659, 11768, 11849, 11950, 12016, 1207, 12144, 12208, 12287, 12360, 12433, 1252, 1262, 12754, 12836, 12936, 10030, 10104, 10292, 10393, 10555, 10625, 10734, 10814, 10890, 10959, 11052, 11149, 11222, 112, 11397, 11481, 11555, 11660, 1176, 1184, 11951, 12017, 12081, 12145, 12209, 12288, 12361, 12434, 12532, 12630, 12755, 12839, 12937, 10031, 10105, 10294, 10394, 10556, 10628, 10737, 10815, 10891, 1095, 11053, 11150, 11223, 11308, 11398, 11482, 11556, 11661, 11771, 11850, 11952, 12018, 12083, 12146, 1220, 1228, 12362, 12436, 12535, 12631, 12758, 1283, 12938, 10032, 10106, 1029, 10395, 10557, 10629, 10738, 10816, 10892, 10960, 11054, 11153, 11224, 11309, 11483, 11557, 11662, 11772, 11851, 11953, 12019, 12084, 12147, 12210, 12291, 12363, 12438, 12536, 12632, 12759, 12841, 12939, 10033, 10107, 102, 10396, 10560, 1062, 10739, 10817, 10893, 10961, 11055, 11156, 11225, 11310, 1139, 11485, 11558, 11664, 11852, 11954, 12020, 12085, 1214, 12292, 12366, 12439, 12537, 12633, 12760, 12843, 12940, 10034, 10108, 10314, 10399, 10561, 10630, 10740, 10818, 10894, 10962, 11056, 11157, 11226, 11311, 113, 11488, 11559, 11665, 1177, 11853, 11955, 12021, 12086, 12150, 12212, 12294, 1236, 12440, 12538, 12634, 12761, 12845, 12941, 10035, 10112, 10317, 1039, 10562, 10631, 10741, 1081, 10895, 10965, 11059, 11158, 11227, 11312, 11400, 11489, 1155, 11666, 1178, 11854, 11956, 12022, 12087, 12151, 12296, 12370, 12441, 12539, 12635, 12762, 12846, 12942, 10036, 10118, 10318, 103, 10563, 10632, 10743, 10826, 10896, 10966, 11060, 11159, 11228, 11313, 11401, 11490, 11565, 11667, 1179, 11855, 11957, 12023, 12088, 12152, 12214, 12297, 12371, 12442, 1253, 12638, 12763, 12847, 12943, 10037, 10119, 10319, 10401, 10564, 10633, 10744, 10897, 10967, 11067, 11160, 11229, 11314, 11402, 11491, 11566, 11668, 117, 11856, 11958, 12024, 12089, 12153, 12215, 12298, 12372, 12443, 12540, 12639, 12764, 12849, 12944, 10038, 10120, 10320, 10402, 10565, 10635, 10747, 10828, 10898, 10968, 11068, 11161, 11230, 11315, 11403, 11493, 11567, 11800, 1185, 12025, 1208, 12154, 12217, 12299, 12373, 12444, 12541, 1263, 12765, 1284, 12945, 10039, 10121, 10321, 10409, 10566, 10636, 10748, 10830, 10899, 10969, 11069, 11162, 11231, 11321, 11408, 11496, 11568, 11671, 11801, 11878, 11960, 12026, 12090, 12155, 12218, 122, 12374, 12445, 12542, 12645, 12766, 12851, 12946, 10040, 10125, 10322, 1040, 10567, 10637, 10760, 10831, 1089, 10970, 11074, 11232, 11322, 11409, 11498, 11569, 11672, 11802, 11880, 11961, 12027, 12091, 12156, 12219, 12300, 12375, 12450, 12545, 1264, 12767, 12852, 12947, 10041, 10127, 10323, 10410, 10568, 10638, 10761, 10832, 108, 10973, 11078, 11164, 11233, 11323, 1140, 11499, 1156, 11673, 11803, 11881, 11962, 12028, 12092, 12157, 12220, 12302, 12376, 12452, 12548, 12652, 12768, 12853, 12948, 10042, 10128, 10324, 10417, 10569, 10639, 10762, 10835, 10900, 10974, 11079, 11165, 11234, 11327, 11410, 114, 11570, 11674, 11882, 11963, 12029, 12093, 12158, 12221, 12304, 12378, 12453, 12549, 12653, 12769, 12854, 1294, 10043, 10129, 10327, 10418, 10570, 1063, 10763, 10837, 10901, 10976, 11080, 11166, 11235, 11328, 11411, 11503, 11571, 11675, 11805, 11883, 11965, 12094, 12159, 12222, 12305, 1237, 12454, 1254, 12654, 12770, 12855, 12951, 10044, 10130, 10328, 1041, 10571, 10642, 10765, 10840, 10902, 10979, 11082, 11167, 11236, 11329, 11416, 11504, 11575, 11676, 11806, 11885, 11966, 12030, 12096, 1215, 12223, 12306, 12380, 12455, 12550, 12655, 12779, 12955, 10045, 10131, 10329, 10422, 10572, 10643, 10766, 10842, 10904, 1097, 11083, 11168, 11237, 11330, 11417, 11505, 11576, 1167, 11807, 11886, 11969, 12031, 12097, 12160, 12224, 12307, 12381, 12456, 12551, 12658, 1277, 1285, 12956, 10046, 10132, 10330, 1043, 10573, 10767, 10843, 10905, 10981, 11084, 11169, 11238, 11331, 11418, 11506, 11577, 1168, 11808, 11887, 1196, 12032, 12098, 12161, 12225, 12308, 12383, 1245, 12552, 12659, 12781, 12860, 12957, 10047, 10133, 10331, 10451, 10576, 10655, 10769, 10844, 10908, 10983, 11085, 11170, 11239, 11332, 11419, 11507, 11578, 11690, 11809, 11888, 11970, 12033, 12099, 12162, 12226, 12309, 12384, 12460, 12553, 1265, 12784, 12863, 12958, 10048, 10134, 10332, 10452, 10577, 10656, 1076, 10847, 10909, 10984, 11088, 11171, 11240, 11333, 11420, 11508, 11579, 11691, 1180, 11889, 11971, 12034, 1209, 12163, 12227, 1230, 12387, 12461, 12554, 12664, 12785, 12864, 1297, 10049, 1013, 10333, 10453, 10578, 10657, 10770, 10848, 1090, 10985, 11089, 11172, 11241, 11334, 11421, 11509, 1157, 11692, 11810, 11890, 11972, 12035, 120, 12164, 12228, 12310, 12388, 12462, 12563, 12667, 12786, 12866, 1298, 10050, 10140, 10335, 10454, 10579, 10658, 10771, 10849, 10910, 10986, 11090, 11173, 11242, 11335, 11422, 1150, 11580, 11693, 11811, 11892, 11973, 12036, 12100, 12166, 12229, 12311, 12389, 12463, 12564, 12668, 12787, 12867, 1299, 10051, 10142, 10338, 10455, 10580, 10659, 10772, 10850, 10911, 10987, 11091, 11174, 11243, 11336, 11424, 11510, 11581, 11694, 11812, 11894, 11974, 12037, 12101, 12169, 1222, 12312, 1238, 12464, 12565, 12669, 12788, 12868, 129, 10052, 10143, 10339, 10458, 10581, 1065, 10773, 10851, 10988, 11092, 11244, 11339, 11426, 11512, 11582, 11695, 11813, 11895, 11975, 12038, 12102, 1216, 12231, 12313, 12390, 12465, 12567, 1266, 1278, 12870, 12, 10053, 10144, 1033, 10459, 10582, 10660, 10776, 10852, 10913, 10989, 11093, 11179, 11245, 11343, 11427, 11513, 1158, 11696, 11814, 11897, 11976, 12039, 12103, 12170, 12232, 12314, 12391, 12466, 12568, 12675, 12792, 12871, 13, 10054, 10147, 10340, 1045, 10583, 10661, 10777, 10853, 10916, 1098, 11095, 11180, 11246, 11344, 1142, 11514, 11599, 11697, 11815, 1189, 11977, 1203, 12104, 12171, 12233, 12392, 12467, 12569, 12676, 12793, 12872, 14, 10055, 10148, 10341, 10460, 10585, 10662, 10778, 10854, 10918, 10991, 11096, 11181, 11247, 11345, 11431, 11515, 1159, 11698, 11816, 118, 11978, 12040, 12106, 12172, 12235, 12317, 12393, 12468, 12570, 12677, 12794, 12874, 15, 10056, 1014, 10342, 10461, 10586, 10665, 10779, 10855, 10919, 10993, 11097, 11182, 11248, 11347, 11433, 11516, 115, 1169, 11817, 11901, 11979, 12041, 12107, 12173, 12236, 12318, 12394, 12469, 12571, 12678, 12797, 12880, 16, 10057, 10159, 10345, 10462, 10587, 10666, 1077, 10856, 1091, 10994, 11100, 11183, 11349, 11435, 11517, 11604, 11707, 11818, 11902, 1197, 12042, 12109, 12175, 12239, 12319, 12395, 1246, 12572, 1267, 12798, 12881, 17, 10058, 1015, 1034, 10463, 10588, 10667, 10780, 10857, 10920, 10995, 11101, 11184, 11250, 11350, 11436, 11518, 11605, 11708, 11819, 11904, 11980, 12043, 1210, 12176, 1223, 1231, 12396, 12470, 12573, 12684, 12799, 12882, 18, 10059, 10161, 10350, 10464, 10589, 10668, 10781, 10858, 10921, 10996, 11102, 11185, 11251, 11351, 11437, 11519, 11606, 11709, 1181, 11906, 11981, 12044, 12110, 12177, 12240, 12320, 12397, 12472, 12574, 12685, 1279, 12885, 19, 1005, 10162, 10351, 1046, 10590, 10670, 10782, 10859, 10922, 10997, 11105, 11186, 11252, 11352, 11438, 1151, 11607, 1170, 11820, 11912, 11982, 12045, 12111, 12179, 12241, 12321, 12398, 12473, 12575, 12686, 127, 12886, 2609, 10060, 10163, 10352, 10472, 10592, 10671, 10783, 1085, 10924, 10998, 11106, 11187, 11253, 11353, 11440, 11520, 11608, 11710, 11821, 11913, 11983, 12046, 12112, 1217, 12242, 12322, 1239, 12474, 12576, 12687, 12800, 12887, 10061, 10164, 10355, 10473, 10593, 10672, 10784, 10860, 10925, 10999, 11107, 11188, 11254, 11354, 11441, 11521, 11609, 11711, 11822, 11915, 11984, 12047, 12114, 12180, 12244, 12323, 123, 12475, 1257, 12689, 12801, 12888, 10062, 10166, 10356, 10474, 10596, 10673, 10785, 10861, 10926, 1099, 11110, 11189, 11255, 11356, 11442, 11524, 11610, 11712, 11823, 11917, 11985, 12048, 12115, 12181, 12245, 12324, 12400, 12476, 12580, 1268, 12803, 1288, 10063, 1016, 10357, 1047, 10597, 10674, 10786, 10862, 10928, 109, 11111, 1118, 11256, 11357, 11448, 11525, 11611, 11713, 11824, 11918, 11986, 12049, 12116, 12183, 12248, 12325, 12402, 12477, 12581, 12690, 12805, 12892}

\begin{figure}[h!]
    \centering
    \includegraphics[width = 0.7 \textwidth]{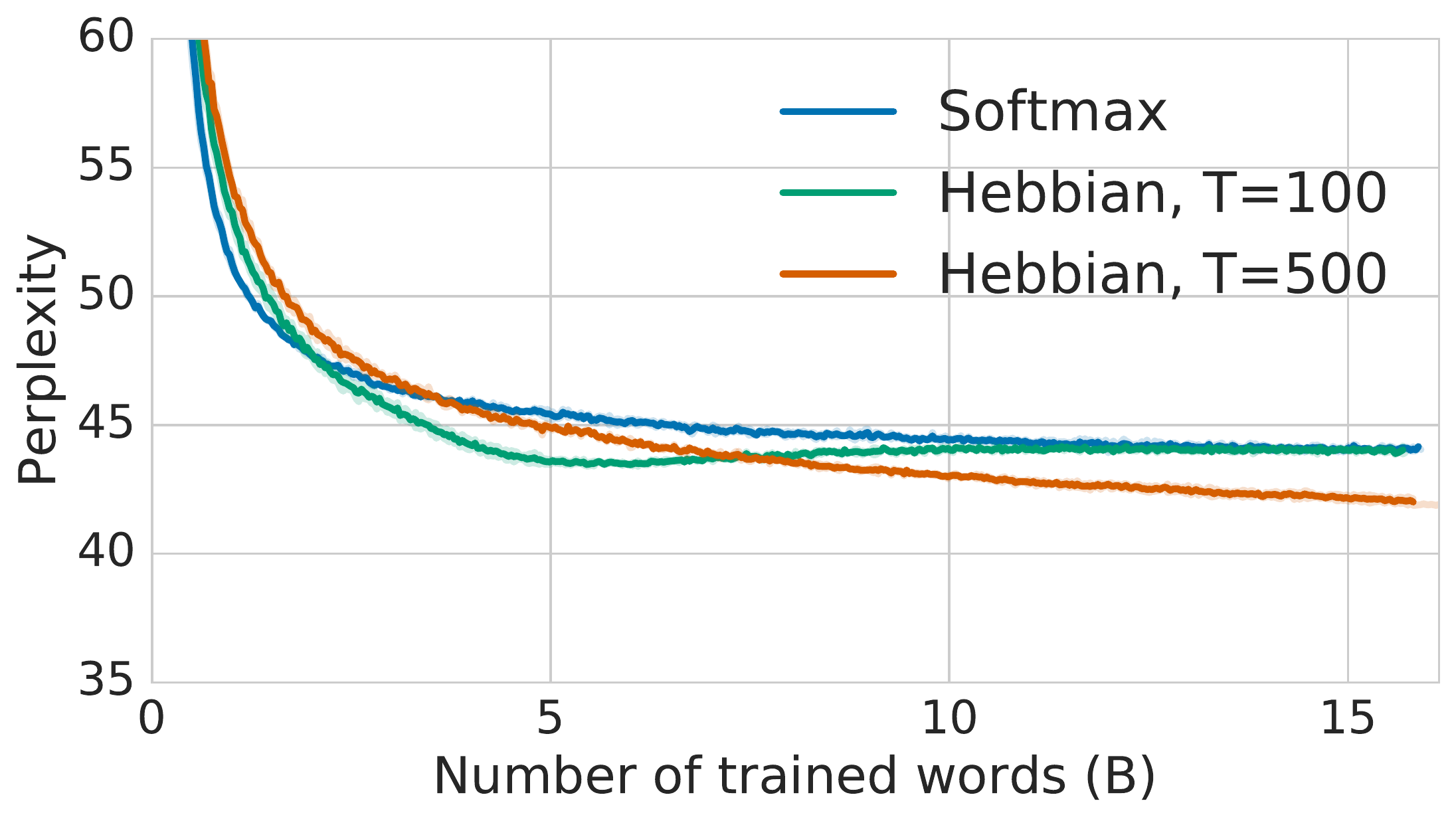}
    \caption{Validation learning curves for the Gutenberg corpus. All word classes have been observed after around $4$B training tokens and we observe the performance of \model return to that of the vanilla LSTM thereafter, as all parameters are optimized by gradient descent.}
    \label{fig:gutenberg_valid}
\end{figure}

\newpage 

\subsection{GigaWord}

\label{appendix:gigaword}
\begin{figure}[h!]
    \centering
    \includegraphics[width=0.7 \textwidth]{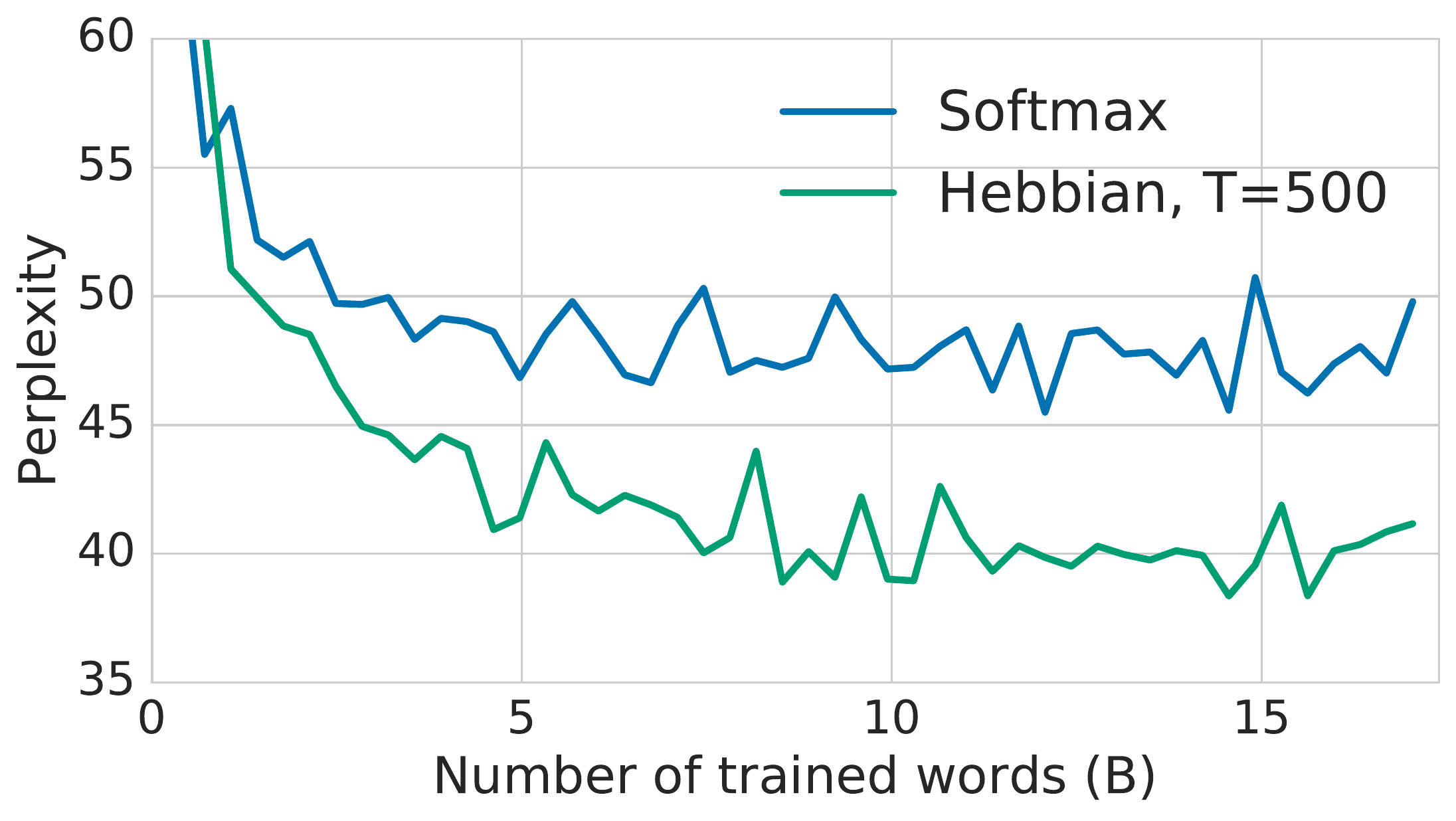}
    \caption{Test perplexity on GigaWord v5 corpus. Each model is trained on all articles from $200-2009$ and tested on $2010$. Because the test set is very large, a random subsample of articles are used per evaluation cycle. For this reason, the measurements are more noisy.}
    \label{fig:gigaword}
\end{figure}

\subsection{Alternate Objective Functions}
\label{appendix:alternates}
\begin{figure}[h!]
   \centering
    \includegraphics[width = 0.7 \textwidth]{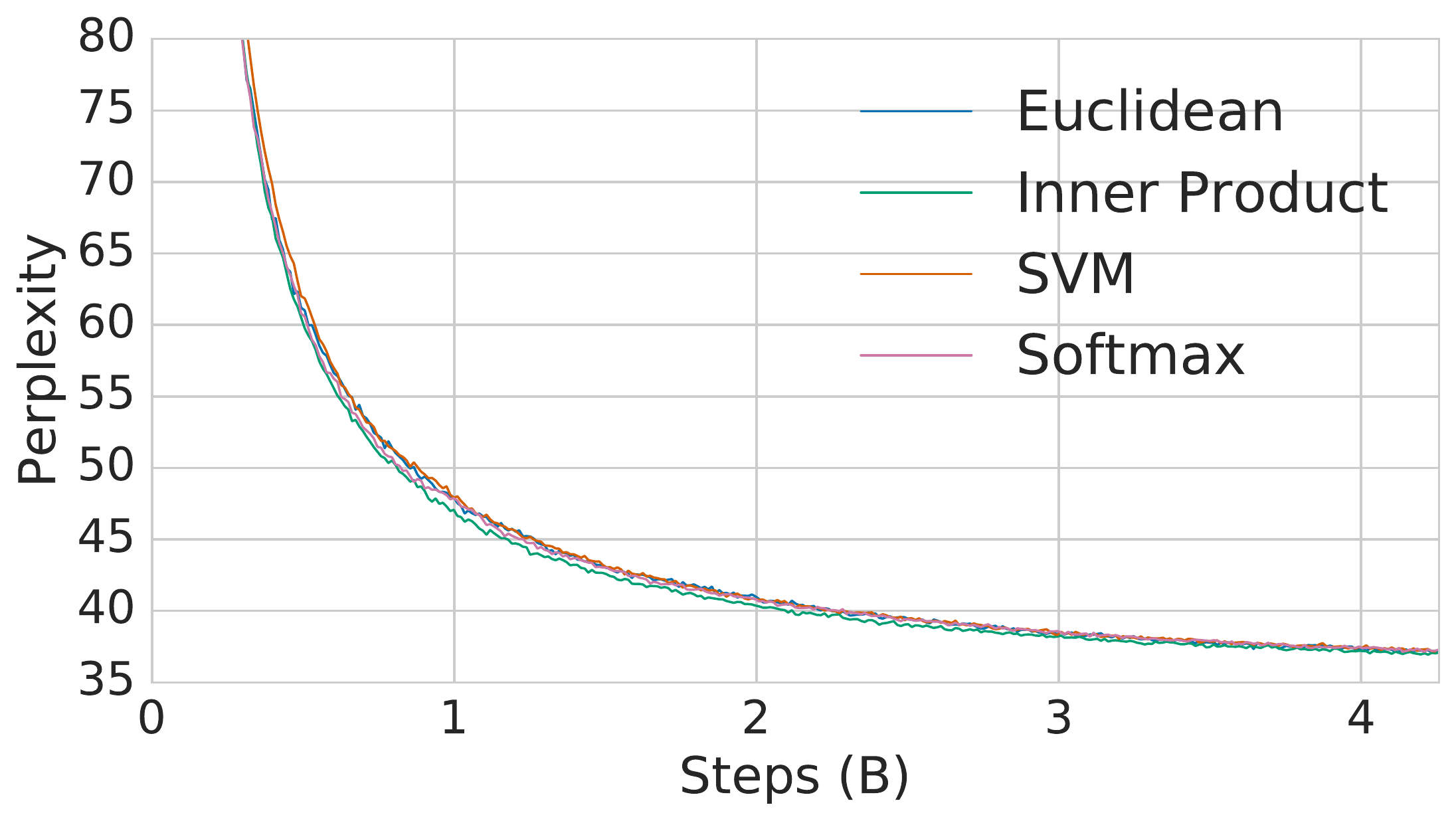}
    \caption{Validation learning curves for WikiText-103 comparing different overfitting objectives. Surprisingly there is not a significant improvement in performance by choosing inner objectives which relate to the overall training objective, e.g. Softmax, vs $L2$. }
    \label{fig:alternates}
\end{figure}

\newpage

\subsection{Data pre-processing}
\label{appendix:preprocess}
For Project Gutenberg and GigaWord v5 we used a very simple python script to pre-process and tokenize the data using NLTK. We post the Gutenberg script here for ease of reproduction. The GigaWord v5 script excludes the Project Gutenberg-specific selection of start / end markers to extract the text. The NLTK library \footnote{\url{https://www.nltk.org/}} is used to split out sentence and word tokens, the resulting text contains lower-case text with one sentence per line.
\begin{small}
\begin{verbatim}
# Copyright 2018 Google LLC.
#
# Licensed under the Apache License, Version 2.0 (the "License");
# you may not use this file except in compliance with the License.
# You may obtain a copy of the License at
#
# https://www.apache.org/licenses/LICENSE-2.0
#
# Unless required by applicable law or agreed to in writing, software
# distributed under the License is distributed on an "AS IS" BASIS,
# WITHOUT WARRANTIES OR CONDITIONS OF ANY KIND, either express or implied.
# See the License for the specific language governing permissions and
# limitations under the License.

def process_text(text):
  import nltk
  start_text = "START OF THIS PROJECT GUTENBERG EBOOK"
  start = text.find(start_text) + len(start_text)
  end = text.find("END OF THIS PROJECT GUTENBERG EBOOK")
  text = text[start:end]
  text = text.decode("utf-8", "ignore")
  text = text.replace("\r", " ")
  text = text.replace("\n", " ")
  final_text_list = []
  sent_text_tokens = nltk.sent_tokenize(text)
  for sentence in sent_text_tokens:
    final_text_list.extend(nltk.word_tokenize(sentence) + ["\n"])
  return " ".join(final_text_list).lower().encode("utf-8")
\end{verbatim}
\end{small}

\newpage

\section{Omniglot}
\label{appendix:omniglot}
\begin{figure}[h!]
    \centering
    \includegraphics[width=0.7 \textwidth]{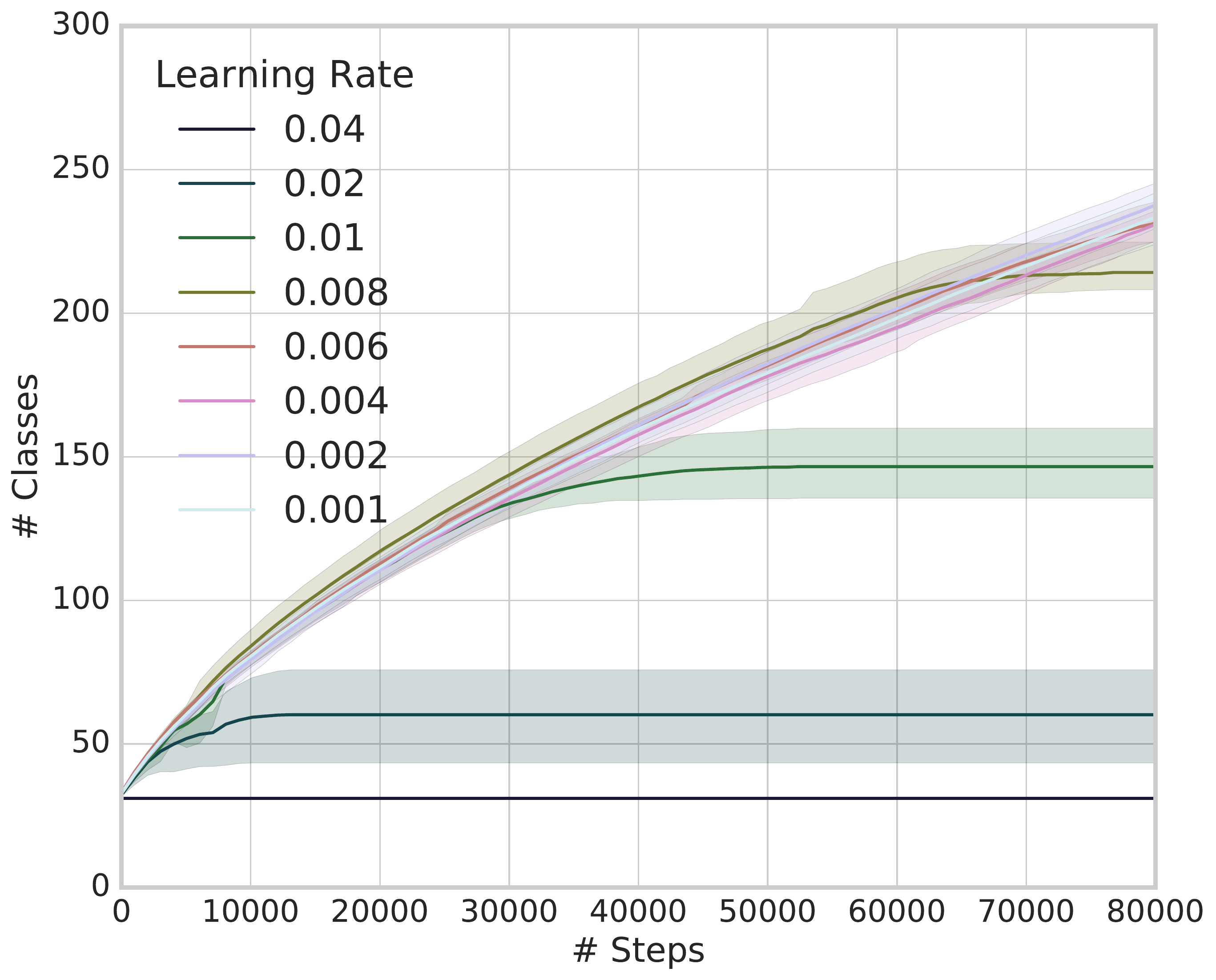}
    \caption{Omniglot curriculum performance versus learning rate for a regular softmax architecture. Values of $1e-3$ to $8e-3$ are similarly fast to learn and are stable. Stability breaks down for larger values.}
    \label{fig:omniglot_lr}
\end{figure}

\begin{figure}[h!]
    \centering
    \includegraphics[width=0.7 \textwidth]{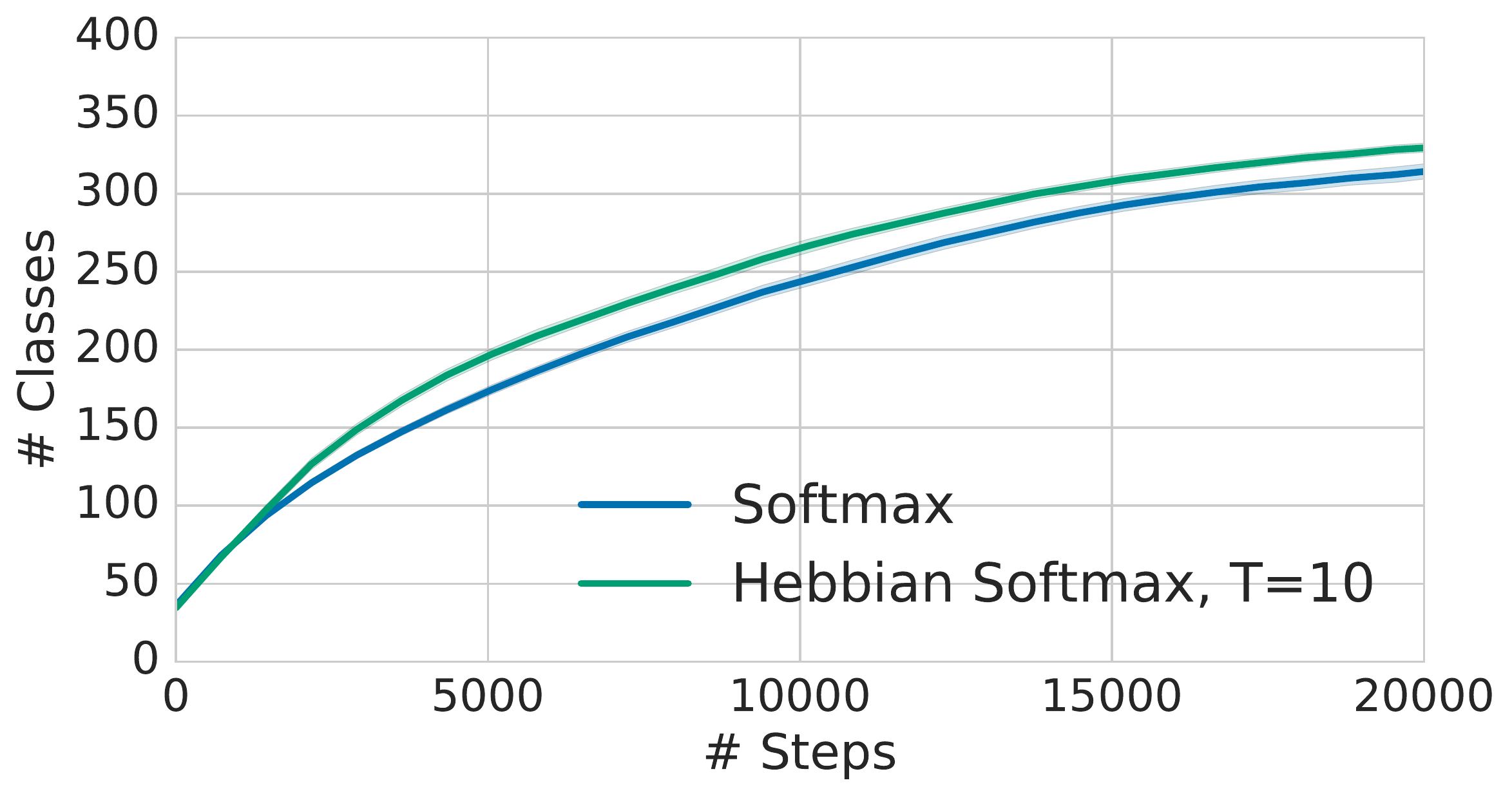}
    \caption{Omniglot curriculum task. Starting from $30$ classes, $5$ new classes are added when total test error exceeds $60$\%. Each line shows a $2\mhyphen\sigma$ confidence band obtained from $10$ independent seed runs.}
    \label{fig:omniglot}
\end{figure}

\end{document}